\documentclass[a4paper,twocolumn,fleqn]{article} 

\usepackage{ist}
\usepackage{xcolor}
\usepackage{gensymb}
\usepackage[normalem]{ulem}
\usepackage{adjustbox}
\usepackage{comment}
\usepackage{amssymb}
\usepackage{amsmath}
\usepackage{graphicx}

\newcommand{\ma}{\textcolor{magenta}}

\newcommand{\ha}{\textcolor{blue}}

\title{Light Direction and Color Estimation from Single Image with Deep Regression}

\author{Hassan A. Sial\textsuperscript{1}, Ramon Baldrich\textsuperscript{1}, Maria Vanrell\textsuperscript{1}, Dimitris Samaras\textsuperscript{2}\\
\textsuperscript{1}Computer Vision Center, Universitat Autònoma de Barcelona, Barcelona, Spain\\
\textsuperscript{2}Department of Computer Science, Stony Brook University, Stony Brook, New York, USA}

\date{} 

\begin{document} 

\maketitle 
\thispagestyle{empty} 


\begin{abstract}
\noindent We present a method to estimate the direction and color of a scene light source from a single image. Our method is based on two main ideas: (a) we use a new synthetic dataset with strong shadow effects with similar constraints to \textit{SID dataset}; (b) we define a deep architecture trained on the mentioned dataset to estimate direction and color of the scene light source. Apart from showing a good performance on synthetic images, we additionally propose a preliminary procedure to obtain light positions of the \textit{Multi-Illumination dataset, and, in this way, we also} prove that our trained model achieves a good performance when it is applied to real scenes.
\end{abstract}

\section{Introduction}

\noindent Scene appearance is directly dependent on the light source properties, such as the spectral composition of emitted light, and the position and direction of the light source, whose interaction with the scene objects provoke shaded surfaces or dark cast shadows that become essential visual cues to understand image content. Esti\-ma\-ting the properties of the light conditions from a single image is an initial step to improve subsequent computer vision algorithms for image understanding. In this paper we perform a preliminary study to estimate color and position of light in a simple and unified approach, that is based on the shading properties of the image where we assume a single scene illuminant.

Estimating the color of the light from a single image has been focus of attention in previous research. Computational color constancy (CC) has been studied in a large number of works \cite{FOSTER2011Color,Finlayson2006, gijsenij2011computational} where the problem was tackled from different points of views \cite{gijsenij2011computational}. A first approach was to extract statistics from RGB image values under different assumptions to estimate the canonical white. A second approach was to introduce spatio-statistical information like gradients or frequency content of the image. One last group of CC algorithms was to try to get the information from physical cues of the image (highlights, shadows, inter-reflections, etc). In the last years, new approaches have been based on deep learning frameworks where the solution is driven by the data with physical constrains in the loss functions. An updated comprehensive compendium and comparison of CC algorithm performances can be found in \cite{cheng2014illuminant,Lou2015DeepColor,das2018color,Yan2018MultipleIE}. In this work we propose one more approach based on a deep architecture, but color of the light source is jointly estimated with the light direction. 

Estimating the direction of the light has also been tackled from different areas like, computer graphics, computer vision or augmented reality. Single image light direction estimation can be divided in two different kinds of approaches.  First, those in which light probes with known reflectance and geometric properties are used. A Specular sphere is commonly used to represent light position in different computer graphics applications   \cite{Debevec1998RenderingSO,Agusanto2003PhotorealisticRF,Schnieders2011LightSE}. But, random shaped objects to detect light position were used by Mandl et-al in \cite{Mandl2017LearningLF}, jointly with a deep learning approach to get the light position with each of these random shaped object. Second, we find those works in which no probe is used, and where multiple image cues such as shading, shape and cast shadows are the basis to estimate light direction. Some examples of these works can be found in computer vision literature \cite{Lalonde2009EstimatingNI,samaras2003incorporating,Panagopoulos2011IlluminationEA,panagopoulos2009robust,arief2012realtime}.

More recently, some deep learning methods have been proposed to estimate scene illumination and have been used for different computer graphics tasks. Gardner et al.\cite{gardner2017learning} introduced a method to convert low dynamic range (LDR) indoor images to high dynamic range (HDR) images, first they used a deep network to localize the light source in LDR image environmental map and then they used another network with these annoted LDR images to convert them to HDR image. Following a similar approach, Geoffroy et al. \cite{hold2017deep} introduced a method to convert outdoor LDR images to HDR images. They trained their network with a set of panorama images and predicted HDR environmental maps with sky and sun positions. Later on, Geoffroy et al \cite{gardner2019deep} extended their previous idea for indoor lighting  but replaced the environmental maps with light geometric and photometric properties. Sun et al. \cite{Sun2019SingleIP} introduced an encoder-decoder based network to relight a single portrait image, the encoder predicts the input image environmental map and an embedding for the scene information, while the decoder builds a new version of the image scene with the target environmental map and obtains a relighted portrait. Very recently, Muramann et al. \cite{murmann2019dataset} have introduced the \textit{Multi-illuminant dataset} of $1000$ scenes each one acquired under $25$ different light position conditions, and they used a deep network to predict a right sided illuminated image from its corresponding left sided illuminated image. In this work we will test our proposal on this new wild dataset after providing a procedure to compute the light direction from each sample.

To sum up, we can state that a large range of works have tackled the problem of estimating color and direction of the scene light source from different points of views and focusing on specific applications. In this work we propose an easy end-to-end approach to jointly characterize the light source of a scene, both for color and direction. We pursuit to measure the level of accuracy we can achieve, in order it can be applied to a wide range of images, without using probes in the scene and becoming a robust preliminary stage to be subsequently combined with any task.  To this end, the paper is organized as follows: first we introduce a new synthetic dataset,  secondly we use it to train a deep architecture that estimates light properties from a single image, finally we show how our proposal performs on three different scenarios, synthetic, real indoor and natural images.

\section{A synthetic dataset}\label{sec:dataset}
\noindent In order to train our end-to-end network that estimates light conditions we developed a new image dataset similar to SID \cite{sial2020deep}, which was created for intrinsic image decomposition. This dataset is formed by a large set of images of surreal scenes where Shapenet objects \cite{shapenet} are enclosed in the center of multi-sided rooms with highly diverse backgrounds provided by flat textures on the room walls, resulting in a large range of different light effects. From now one we will refer to this dataset as SID1, and we will propose a new one better adapted to our problem, using the same methodology and software provided by the authors, which is based on an open source Blender rendering engine to synthesize images.

Our new dataset will be called SID2 dataset. The main difference is that it introduces more than one object in each scene, with the aim of increasing the number of strong light effects and interactions. Additionally, we also introduce more variability in the distance from the light source to the scene center. The dataset is formed by $45,000$ synthetic images with the corresponding ground truth data: direction and color of the light source.

\begin{figure}
\begin{tabular}{c}
 \includegraphics[width=0.45\textwidth]{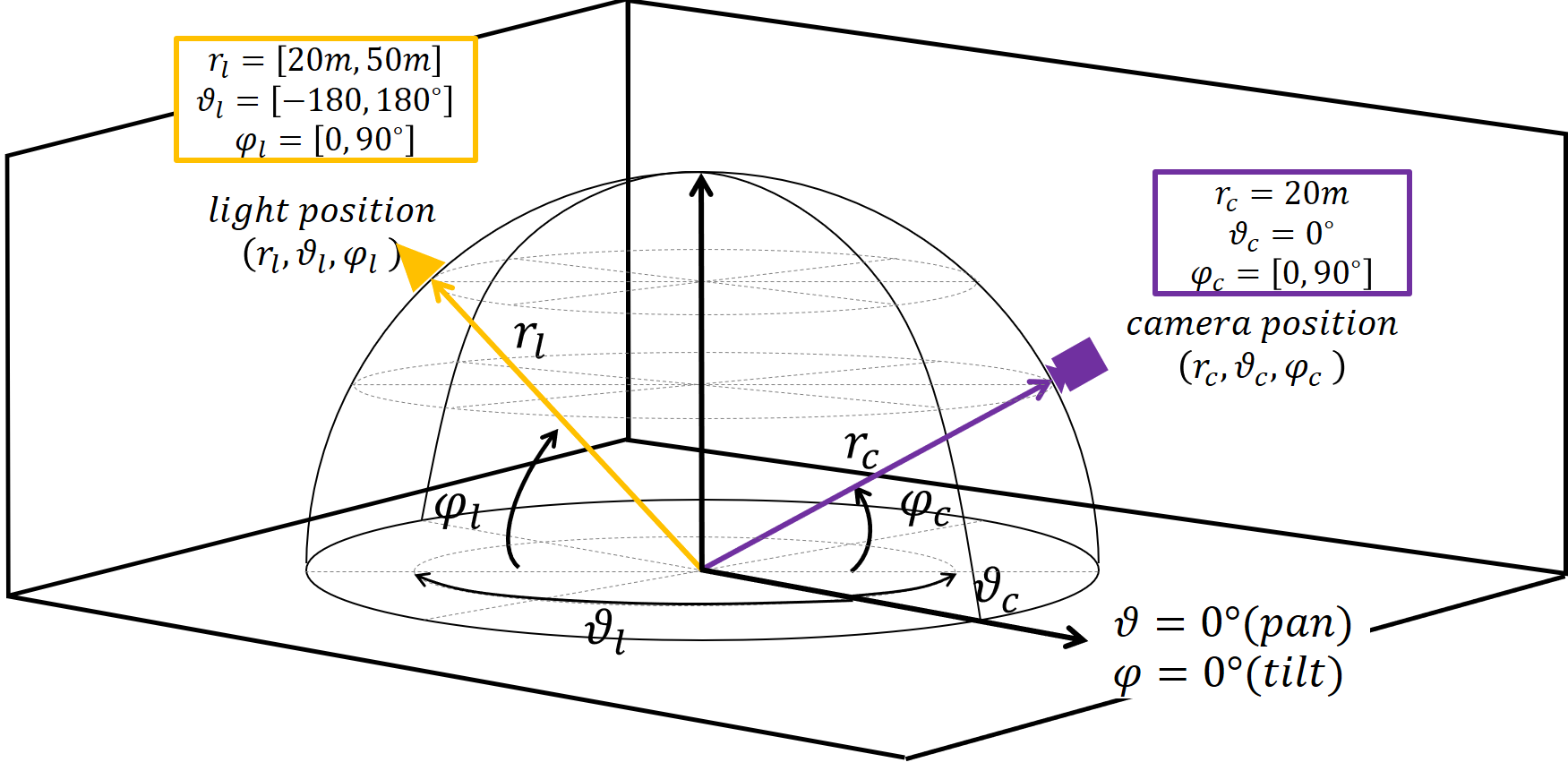}
 \\
\end{tabular}
  \caption{Image Generation Setup. Camera and light positions are given in spherical coordinates $(r, \vartheta,\varphi)$.}
 \label{Figure:generation_setup}
\end{figure}

We did several assumptions in building the dataset: (a) objects are randomly positioned around the scene center but always close to the room ground floor to have realistic cast shadows; (b) light source direction goes from scene center to a point onto an imaginary semi-sphere of a random radius and with random RGB color; (c) camera is randomly positioned at a random distance from the center of the scene and always with the focal axis pointing to the scene center. In Figure \ref{Figure:generation_setup} we show the  diagram of the synthetic world we defined for the generation of SID2. More specifically, we took $45,000$ 3D objects from Shapenet dataset \cite{shapenet}. Likewise in \cite{sial2020deep} we did not use textured surfaces, we used a diffuse bidirectional scattering distribution function (BSDF) with random color and roughness values for each mesh texture in each object. This roughness parameter controls how much light is reflected back from each object surface. We randomly picked from $1$ to $3$ objects in each image. They were placed at random locations within the camera view range. We placed an empty object in the center of the scene to ensure non-overlapping between the rest of objects. Light direction was randomly defined in spherical coordinates $(r_l,\vartheta_l,\varphi_l)$, being radius, pan and tilt, respectively. We took random values within the ranges of $[20m,50m]$ $[30^{\circ},90^{\circ}]$ and $[0^{\circ},360^{\circ}]$ respectively, in steps of $1^{\circ}$ for pan and $5^{\circ}$ for tilt. Light intensity and chromaticity was randomly selected, but chromaticity was constrained to be around the Planckian locus to simulate natural lighting conditions. Camera position is also denoted in spherical coordinates as $(r_c,\vartheta_c,\varphi_c)$, where $r_c$ was fixed at $20m$ and pan, $\vartheta_c=0^{\circ}$, the tilt range randomly varied within $[10^{\circ},70^{\circ}]$. In the final ground-truth (GT), light pan and tilt are provided with reference to the camera position, in order not to depend on real world positions which are usually not available in real images. Backgrounds were generated in the same way as in \cite{sial2020deep}.

\section{Our deep architecture}
\begin{figure}
    \centering
    \includegraphics[width=0.48\textwidth]{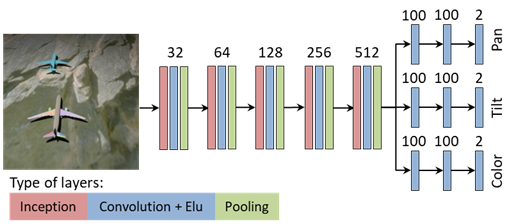}
    \caption{Deep Architecture. Inception module from \cite{szegedy2016rethinking}}
    \label{fig:deep architecture}
\end{figure}

\noindent We propose an inception-based encoder-decoder architecture to predict light parameters. In Figure \ref{fig:deep architecture} we give an scheme, where we can see that our encoder has five modules combining 3 types of layers: inception, convolution and pooling. The encoder input is the image that is transformed to a higher dimensional feature space, from which three decoders convert this embedding to a common feature space of pan, tilt and color of light source.  Pan and tilt output predictions are given as functions of angle differences. We use the functions  $\sin(\vartheta_c -\vartheta_l)$ and $\cos(\vartheta_c -\vartheta_l)$ to bound the pan output. Similarly, tilt prediction is represented as difference of angles $\sin(\varphi_c - \varphi_l)$ and $\cos(\varphi_c - \varphi_l)$. Finally color is predicted here as R, G and B values. We used the split inception module from \cite{szegedy2016rethinking}, which replaces $n \times n$ convolution filters with $1 \times n$ and $n \times 1$ filter, to achieve faster convergence with overall less parameter. Our global loss function to estimate illumination parameters is based on three terms:
\begin{equation}  Loss(x,\hat{y}) = 
\alpha_1 L_{Pan}(x,\hat{y}) +
\alpha_2 L_{Tilt}(x,\hat{y}) +
\alpha_3 L_{Color}(x,\hat{y}) \end{equation}
where $x$ is the input image, $\hat{y}$ is a 7 dimensional vector giving the estimation of the scene light properties represented by $x$, $\alpha_i$ are the weights for the different loss terms defined for pan, tilt and color, and which are respectively given by:
\begin{equation}
\begin{array}{llll}
L_{Pan}(x,\hat{y}) = MSE((\hat{y}_1 - \sin(\vartheta_c^x -\vartheta_l^x)) + (\hat{y}_2 - \cos(\vartheta_c^x -\vartheta_l^x))) &  \\
       L_{Tilt}(x,\hat{y}) = MSE\{(\hat{y}_3 - \sin(\varphi_c^x -\varphi_l^x)) + (\hat{y}_4 - \cos(\varphi_c^x -\varphi_l^x)\} 
     & \\ L_{Color}(x,\hat{y}) = \arccos((\hat{y}_5 \cdot x_{RGB}) / \|\hat{y}_5\| * \|(x_{RGB}\|) 
\end{array}\nonumber 
\end{equation} 
$L_{Pan}$ and $L_{Tilt}$ are computed as the mean square error ($MSE$) between the estimations for pan, $\hat{y}_1$ and $\hat{y}_2$, and for tilt, $\hat{y}_3$ and $\hat{y}_4$, and a function of the difference between the camera and light positions for the ground-truth of $x$. The third loss term, $L_{Color}$, is the mean angular error between the estimated RGB values, $\hat{y}_5$, and the color of the light for $x$ image provided in the ground-truth.

This network has been trained using Adam optimizer \cite{AdamKingmaB14}, with initial learning rate $0.0002$ which is decreased with factor of $0.1$ on reaching plateau. Weights are initialized using He Normal\cite{He2016DeepRL}. All experiments in next sections were trained using a batch size of $16$. In the following sections we show the results of several experiments to evaluate the architecture performance on different datasets and conditions.

\section{Experiment 1: Synthetic dataset}
\noindent In this first experiment we trained and tested the proposed architecture on two different datasets: SID1 (single object) and SID2 (multiple objects). The results are shown in Table \ref{tab:Exp_SID}, where we separately compute different angular errors. Direction error is given separately in the pan and tilt components, and the global angular error for direction estimation. We can see that all the estimations are improved when the network is trained on a more complex dataset, like SID2, where multiple objects light interactions provide richer shading cues. However, there is slight improvement for color estimation, performance is similar for both datasets. In Figure \ref{Figure:SID_examples} we show qualitative results on SID2 dataset. Images are ordered from smaller (left) to larger (right) direction estimation error.

\begin{table}[h!]
\tiny
\centering
\resizebox{\columnwidth}{!}{\begin{tabular}{|c|c|c|c|c|} 
\hline
Dataset & Pan & Tilt & Direction & Color \\
    \hline \hline
    SID1 & $14.63$ & $9.86$ & $16.98$ & $1.05$\\ \hline
    SID2 & $10.46$ & $9.21$ & $14.22$ & $1.02$ \\ \hline
    \end{tabular}}
    \caption{\textit{Table 1.} Estimation Errors (in degrees) for light source direction and color with the proposed architecture trained on SID1 and SID2.}
    \label{tab:Exp_SID}
\vspace{4mm}
\end{table}

Intuitively as the light becomes more zenithal, the shadows shorten and cast shadows present more uncertainty to estimate light direction. We analysed the performance of the method at different tilt locations of the light source in the input image, from the ground (level 1: $[30^\circ$, $50^\circ]$) up to the zenithal area (level 3: $[70^\circ$, $90^\circ]$). This effect is confirmed in Table \ref{tab:my_label2},  where estimated errors in direction clearly increase from level 1 to level 3. 

\begin{table}[!h]
\tiny
\centering
    \resizebox{\columnwidth}{!}{\begin{tabular}{|c|c|c|c|c|}
    \hline
     Tilt range  & Pan  & Tilt & Direction & Color \\
    \hline \hline
    Level 1 & $4.92$ & $5.14$ & $7.57$ & $1.04$\\ \hline
    Level 2 & $8.51$ & $5.14$ &	$11.97$ & $0.90$ \\ \hline
    Level 3 & $22.36$ & $18.33$ & $28.33$ & $1.00$\\ \hline
    \end{tabular}}
    \caption{\textit{Table 2.} Estimation Errors (in degrees) for light direction and color at different tilt levels.}
    \label{tab:my_label2}
\vspace{4mm}
\end{table}

Similarly, we analysed the performance at different pan levels, each level covers $90\degree$ of pan area. Level 1 is when light comes from center front, level 2 from right, level 3 from back and level 4 is when light comes from left side. Tilt angle was kept between $30\degree$ and $70\degree$ to analyze pan error while minimizing zenithal tilt error effects. Table \ref{tab:my_label3} shows results for this experiment, both direction and color error are consistent in all levels of pan.

\begin{table}[!h]
    \centering
    \tiny
    \resizebox{\columnwidth}{!}{\begin{tabular}{|c|c|c|c|c|}
    \hline
       Pan range  & Pan  & Tilt & Direction & Color \\ 
    \hline \hline
    Level1 & $6.87$ & $5.10$ & $9.10$ & $0.98$\\ \hline
    Level2 & $6.24$ & $5.12$ &	$8.80$ & $0.96$ \\ \hline
    Level3 & $6.35$ & $5.09$ & $9.46$ & $1.03$\\ \hline
    Level4 & $6.01$ & $5.16$ & $9.03$ & $0.97$\\ \hline
    \end{tabular}}
    \caption{\textit{Table 3.} Estimation Errors (in degrees) for light direction and color at different pan levels.}
    \label{tab:my_label3}
\vspace{0.4cm}
\end{table}

\section{Experiment 2. Multi-illumination dataset}
\noindent Once we have evaluated our  method in synthetic images, we want to analyze whether it generalises for real images. We have tested our method on the \textit{Multi-illuminant dataset} (MID)  \cite{murmann2019dataset}, that contains $1000$ different indoor scenes, all of them containing a diffuse and a specular sphere at random locations. Light source is mounted above the camera and can be rotated at different predefined pan and tilt angles, creating different light conditions. The dataset provides the orientation of the light source for each acquired image, but since the light can 
bounce off the walls, the direction of the incident light on the scene is not defined by the light source angles and it needs to be recomputed.

We have defined a procedure to compute the incident light direction from the specular sphere present in all the scenes, whose highlights provide enough information to collect our GT data (tilt and pan angle between light and camera). The color of the light is obtained from the average color of the diffuse sphere. To obtain the light direction we used the ideas proposed by \cite{Schnieders2011LightSE} where they assume that the angle of incident light is equal to the angle of outgoing light at the specular highlight on a spherical ball. We use a reference image in each scene where light and camera both are pointed in the same direction towards the center of the scene. We also assumed that the light is mounted at $10\degree$ height with respect to scene center. The angles obtained from this reference images allow to correct the angle displacement due to the sphere position shifting inside the image on the rest of scene images.
\begin{table}[!h]
    \centering
    \small
    \resizebox{\columnwidth}{!}{\begin{tabular}{|l|c|c|c|c|}
    \hline
      Dataset (Error) & Pan  & Tilt & Direction & Color \\ \hline \hline
    Masked MID (Mean) & $21.38$ & $10.14$ & $22.72$ & $0.63$ \\ 
    Masked MID (Median) & $13.74$ &	$7.64$ & $17.80$ & $0.40$  \\ \hline
    UnMasked MID (Mean) & $14.28$ & $6.96$ & $15.44$ & $0.36$ \\ 
    UnMasked MID (Median) & $8.20$ & $4.83$ & $10.83$ &	$0.24$  \\ \hline 
    \end{tabular}}
    \caption{\textit{Table 4.} Estimation errors in degrees on two versions of MID dataset (with Masked or UnMasked spheres).}
    \label{tab:wild1}
\vspace{0.4cm}
\end{table}

Starting from the network trained on SID2 it was fine tuned on this dataset under two different conditions: a) keeping the reference spheres in the image, and b) masking them. Although specular spheres are not present in real images, the first configuration should provide an upper bound of our method performance on wild images. Table \ref{tab:wild1} shows the results on this experiment. As expected, network performance is much higher when complete images are used as inputs. We can also observe that results on color estimation are better than on SID2, mainly due to the stability of single white light source in the dataset.  To analyze the results removing the influence of the outliers,  we also reported median error on this dataset. Results are as good as the ones obtained only using synthetic images on the upper bound. Qualitative results are provided in Figure \ref{Figure:wild_examples}. Top row depicts the original image, second row are the spheres generated from the GT information, and the third row shows the synthetic spheres generated with the obtained prediction.

Finally, we perform a last experiment on this dataset by dividing the test set in two: (a) images with incident light from the front, and (b) from the back. Table \ref{tab:wild2} also shows the errors computed for these two sets. Both color and direction errors are higher when the light comes from the back of the scene and a big area of the image becomes saturated. We want to note here that the GT we created present a low accuracy for the subset of images with back light sources.This is due to the inherent uncertainty derived from what can be inferred from spheres illuminated from the back. Therefore, this MID dataset division is highly recommended to analyse results derived from this GT.

\begin{table}[!h]
    \centering
    \tiny
    \resizebox{\columnwidth}{!}{\begin{tabular}{|c|c|c|c|c|}
    \hline
   Light position  & Pan  & Tilt & Direction & Color\\
    \hline \hline
    Front & $11.51$ & $6.33$ &	$13.09$ & $0.34$
\\ \hline
    Back & $32.90$ & $11.21$ & $31.23$ & $0.52$
 \\ \hline 
    \end{tabular}}
    \caption{\textit{Table 5.} Estimation errors in degrees dividing MID dataset in front and back light.}
    \label{tab:wild2}
\vspace{0.4cm}
\end{table}

\renewcommand{\arraystretch}{1.5}

\newcommand\sizephotosyn{0.15}
\begin{figure*}[h!]
\begin{center}
\setlength{\tabcolsep}{1pt}
\begin{tabular}{rcccccc}
   \raisebox{1.2cm}{(a)}  & \includegraphics[width=\sizephotosyn\textwidth]{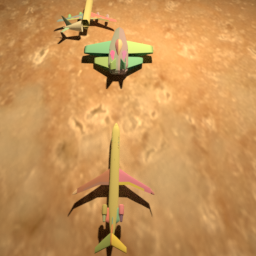} &
     \includegraphics[width=\sizephotosyn\textwidth]{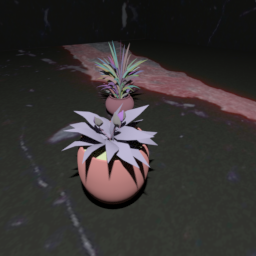} &
      \includegraphics[width=\sizephotosyn\textwidth]{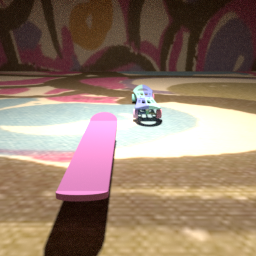} &
       \includegraphics[width=\sizephotosyn\textwidth]{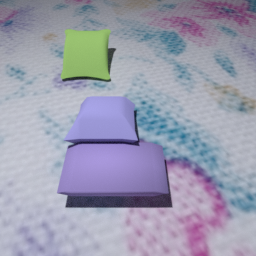} &
        \includegraphics[width=\sizephotosyn\textwidth]{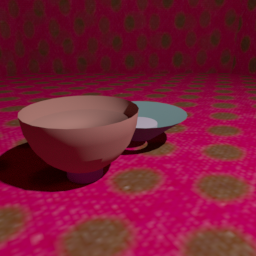} &
         \includegraphics[width=\sizephotosyn\textwidth]{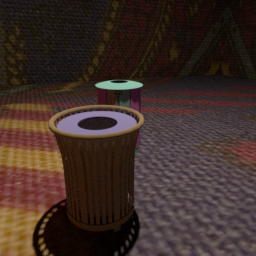} \\[-0.1cm]
   \raisebox{1.2cm}{(b)} & \includegraphics[width=\sizephotosyn\textwidth]{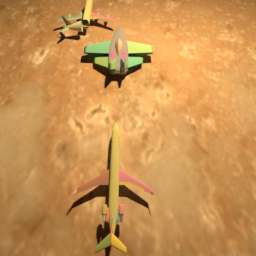} &
     \includegraphics[width=\sizephotosyn\textwidth]{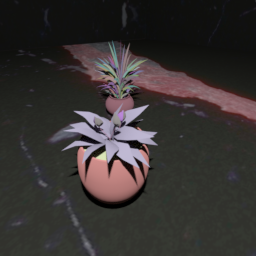} &
      \includegraphics[width=\sizephotosyn\textwidth]{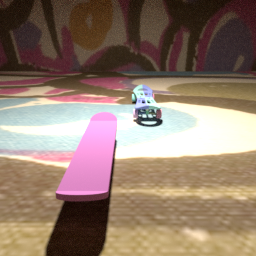} &
       \includegraphics[width=\sizephotosyn\textwidth]{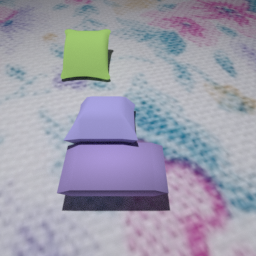} &
        \includegraphics[width=\sizephotosyn\textwidth]{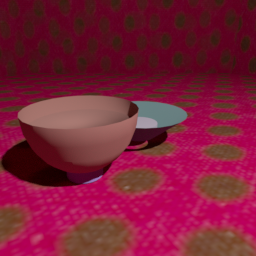} &
         \includegraphics[width=\sizephotosyn\textwidth]{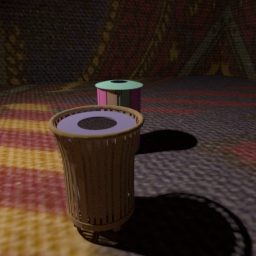} \\[-0.1cm]
     \raisebox{1.2cm}{(c)}& \includegraphics[width=\sizephotosyn\textwidth]{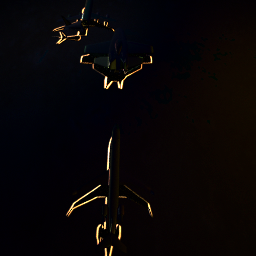} &
     \includegraphics[width=\sizephotosyn\textwidth]{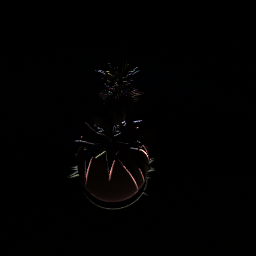} &
      \includegraphics[width=\sizephotosyn\textwidth]{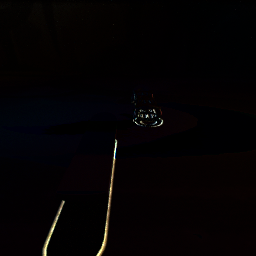} &
       \includegraphics[width=\sizephotosyn\textwidth]{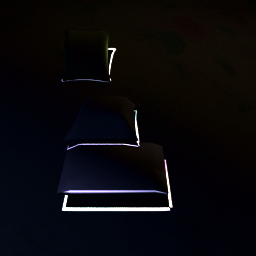} &
        \includegraphics[width=\sizephotosyn\textwidth]{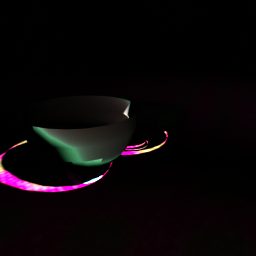} &
         \includegraphics[width=\sizephotosyn\textwidth]{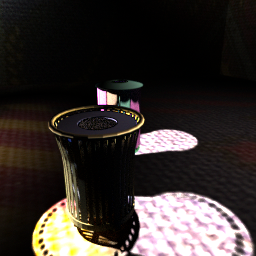} \\[-0.1cm] \hline\hline
                 $Direction$ & $0.57$ & 1.58 & $4.3$ &  $8.55$ & $8.81$ & $30.0$  \\[-0.2cm]
          $Color$ & $1.27$ & $0.78$ & $1.64$ & $1.8$ & $0.22$ & $0.24$  \\ \hline
         
\end{tabular}
\end{center}
\caption{Direction and Color estimation examples on SID2 dataset: (a) Original images, (b) Generated images with estimated light properties, (c) RGB Image subtraction between (a) and (b). Bottom rows are the corresponding computed errors for direction and color in degrees, ordered from smaller (left) to larger (right) direction estimation error.} \label{Figure:SID_examples}
\end{figure*}
\newcommand\sizephoto{0.225}
\begin{figure*}[h!]
\begin{center}
\setlength{\tabcolsep}{1pt}
\begin{tabular}{rcccc}

   \raisebox{1.0cm}{(a)}  & \includegraphics[width=\sizephoto\textwidth]{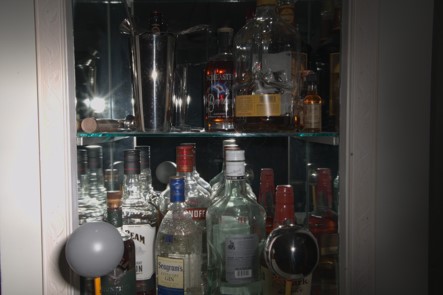} &
  \includegraphics[width=\sizephoto\textwidth]{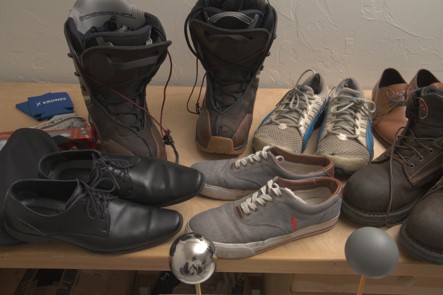} &
  \includegraphics[width=\sizephoto\textwidth]{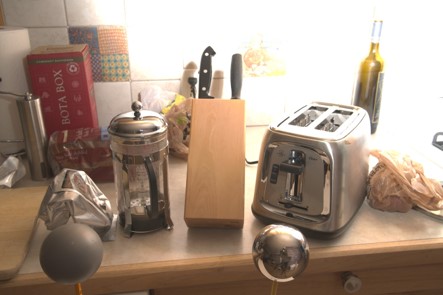} &
  \includegraphics[width=\sizephoto\textwidth]{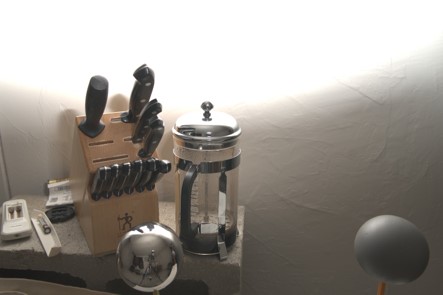}\\[-0.1cm]
   \raisebox{0.3cm}{(b)} &
   \includegraphics[width=\sizephoto\textwidth]{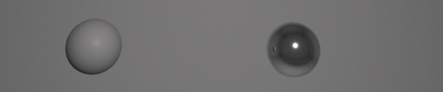} &
  \includegraphics[width=\sizephoto\textwidth]{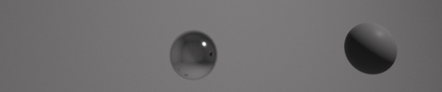} &
  \includegraphics[width=\sizephoto\textwidth]{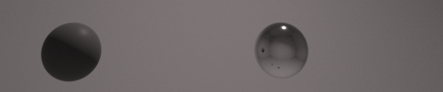} &
  \includegraphics[width=\sizephoto\textwidth]{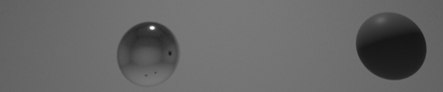}\\[-0.1cm]
     \raisebox{0.3cm}{(c)}&    \includegraphics[width=\sizephoto\textwidth]{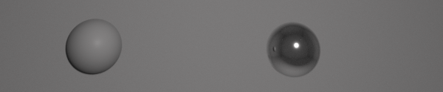} &
  \includegraphics[width=\sizephoto\textwidth]{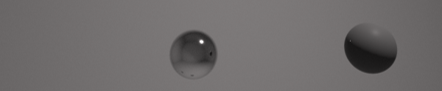} &
  \includegraphics[width=\sizephoto\textwidth]{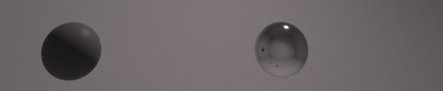} &
  \includegraphics[width=\sizephoto\textwidth]{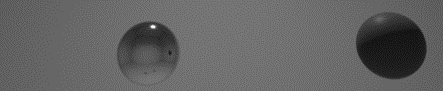}\\ [-0.1cm]\hline\hline
          $Direction$ & $3.16$ & $5.60$ & $20.12$ &  $25.05$   \\[-0.2cm]
          $Color$ & $0.16$ & $0.21$ & $1.30$ & $0.35$  \\ \hline
\end{tabular}
\end{center}
\caption{Direction and Color estimation examples on Multi-illumination dataset: (a) Original images, (b) Ground-truth plotted on corresponding spheres, (c) Estimations provided by our proposed architecture. Bottom rows are computed errors for direction and color in degrees.} \label{Figure:wild_examples}
\end{figure*}

\section{Experiment 3. Natural images}
\noindent Previous experiments show the performance of our method on synthetic and real indoor images. Here, we show a few  qualitative results on real outdoor images. In Figure \ref{Figure:real_examples} we show some examples with strong outdoor cast shadows, in order to visually evaluate the prediction we depict a synthetic pole at the left top corner. In these examples camera is assumed to be at $45\degree$ tilt from ground. Left side four images are from SBU shadow dataset \cite{vicente2016large} and the two on the right have been captured with a mobile device.

\newcommand\sizephotosynb{0.16}
\begin{figure*}[h!]
\begin{center}
\setlength{\tabcolsep}{1pt}
\begin{tabular}{cccccc}
    \includegraphics[width=\sizephotosynb\textwidth]{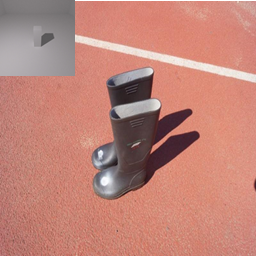} &
    \includegraphics[width=\sizephotosynb\textwidth]{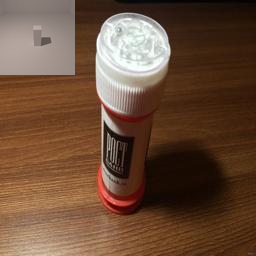} &
    \includegraphics[width=\sizephotosynb\textwidth]{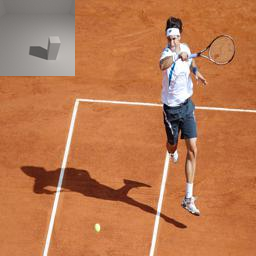} &
    \includegraphics[width=\sizephotosynb\textwidth]{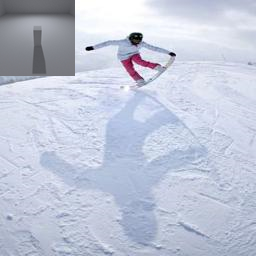} &
    \includegraphics[width=\sizephotosynb\textwidth]{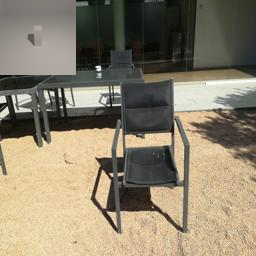} &
    \includegraphics[width=\sizephotosynb\textwidth]{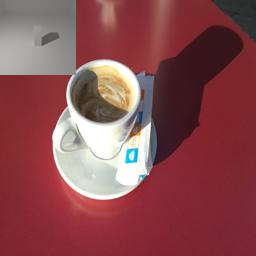} \\[-0.1cm]
\end{tabular}
\end{center}
\caption{Examples of light direction estimation on natural images. Predicted direction is plotted top left in each image.}
\label{Figure:real_examples}
\end{figure*}

\section{Conclusions}

In this work we have proved the plausibility of using a simple deep architecture to estimate physical light properties of a scene from a single image. The proposed approach is based on training a deep regression architecture on a large synthetic and diversified dataset. We show that the obtained regressor can generalize to real images and can be used as a preliminary step for further complex tasks. 

\bibliographystyle{plain}
\small
\bibliography{biblio}








\end{document}